\documentclass[conference]{IEEEtran}
\IEEEoverridecommandlockouts
\usepackage{cite}
\usepackage{amsmath,amssymb,amsfonts}
\usepackage{algorithmic}
\usepackage{graphicx}
\usepackage{textcomp}
\usepackage{xcolor}
\usepackage{balance}
\usepackage{soul}
\def\BibTeX{{\rm B\kern-.05em{\sc i\kern-.025em b}\kern-.08em
    T\kern-.1667em\lower.7ex\hbox{E}\kern-.125emX}}
\begin{document}

\title{Constrained Bayesian Optimization for Hierarchical Federated Learning in IoT Networks for Plant Disease Classification

}

\author{Athanasios~Papanikolaou\textsuperscript{\ddag},~Athanasios~Tziouvaras\textsuperscript{†},~Apostolos~Xenakis\textsuperscript{†}, 
Periklis~Chatzimisios\IEEEauthorrefmark{6},\\~Shameem~A.~Puthiya Parambath\textsuperscript{+}, George~Floros\textsuperscript{\S}, Enrica~Zereik\textsuperscript{*}, Ivan~Petrovic\textsuperscript{\ddag},  and~Fabio~Bonsignorio\textsuperscript{\ddag}\\
\textsuperscript{\ddag}University of Zagreb, Croatia, emails:
\{athanasios.papanikolaou, ivan.petrovic, fabio.bonsignorio\}@fer.unizg.hr\\
\textsuperscript{†}University of Thessaly, Greece, emails:
\{attziouv, axenakis\}@uth.gr\\
\IEEEauthorrefmark{6}International Hellenic University, email:
pchatzimisios@ihu.gr\\
\textsuperscript{+}University of Glasgow, UK, email:
sham.puthiya@glasgow.ac.uk\\
\textsuperscript{\S}Trinity College Dublin, Ireland, email:
florosg@tcd.ie\\
\textsuperscript{*}Italian National Research Council, Institute of Marine Engineering, Italy, email:
enrica.zereik@cnr.it
}

\maketitle

\begin{abstract}
The deployment of Hierarchical Federated Learning (HFL) in resource-constrained Internet of Things (IoT) environments requires careful configuration to balance predictive performance with energy consumption and execution time. This challenge is particularly relevant to smart agriculture, where distributed IoT devices can support automated plant disease classification while operating under limited computational and communication resources. This paper presents a constrained Bayesian Optimization framework for the efficient configuration of HFL deployments. The proposed approach jointly explores the deep learning backbone architecture, aggregation strategy, and number of communication rounds, while the federation size is determined according to the spatial coverage requirements of the agricultural deployment. A weighted objective function captures user-defined trade-offs among energy consumption, execution time, and predictive performance, while explicit constraints ensure compliance with deployment-specific resource and accuracy requirements. The framework is evaluated on an IoT-based plant disease classification task considering multiple deep learning architectures, federated aggregation strategies, and communication-round settings. Experimental results across 30 independent optimization runs show that the proposed approach explores only 11.11\% of the search space, while consistently identifying solutions within 1\% of the exhaustive-search optimum, with a mean optimality gap of only 0.056\%.
\end{abstract}


\section{Introduction}
The agricultural sector is undergoing a major transformation, which is mainly driven by the combination of Internet of Things (IoT) technologies and deep learning (DL) models. This shift has already created new application scenarios such as data-driven crop monitoring, early disease detection and precision resource management \cite{wolfert2017bigdata, elijah2018iot}. Within this context, the identification of plant diseases remains a critical challenge that undermines food security, since delayed interventions may result in substantial yield losses and economic damage \cite{vishnoi2021plant}. A possible solution could reside within the recent advances in computer vision and convolutional neural networks, which have achieved high accuracy in automated plant disease classification \cite{upadhyay2025deep, delnevo2021deep}. However, the deployment of centralized DL pipelines in large-scale agricultural environments raises practical concerns. This mainly happens because continuous data transmission to remote cloud servers is impractical for resource-constrained IoT networks that operate under limited bandwidth, energy, and computational budgets \cite{kashyap2021towards}.

Federated Learning (FL) has emerged as a compelling distributed paradigm that enables multiple IoT nodes to collaboratively train a shared model without exchanging raw data. Thus, this approach preserves data privacy and reduces communication overhead, compared with other distributed deployments \cite{beltran2023decentralized, mcmahan2017fedavg}. Several recent works have explored FL for agricultural applications, including crop disease classification \cite{hari2025adaptive, behera2025crop} and yield prediction \cite{bera2024flag}. These works demonstrate that FL can achieve competitive performance, while respecting the privacy and autonomy of individual farm sites. Despite these advances, the practical deployment of FL on heterogeneous IoT devices remains challenging, as system performance is highly sensitive to several configuration parameters. Such parameters include but are not limited to the backbone DL architecture, the model aggregation strategy (e.g., FedAvg \cite{mcmahan2017fedavg}, FedProx \cite{li2020fedprox}, FedAvgM \cite{hsu2019fedavgm}), the number of communication rounds and the number of participating devices \cite{wang2019adaptive}. In the majority of the existing literature, these design choices are selected through manual experimentation or via an exhaustive search. Unfortunately, both strategies are prohibitively expensive when each configuration evaluation requires end-to-end execution of the full FL pipeline.

In this paper, we extend  \cite{papanikolaou2025distributed, 11564301} by introducing a constrained Bayesian Optimization framework for resource-aware Hierarchical Federated Learning (HFL) configuration in IoT-based plant disease classification. The proposed framework systematically explores deployment-specific configurations while accounting for resource and predictive-performance requirements. The main contributions of this work are summarized as follows:

\begin{itemize}
\item We formulate HFL configuration as a constrained optimization problem that jointly considers architectural and training parameters under deployment-specific requirements.
\item We introduce a configurable weighted objective function that captures user-defined trade-offs among energy consumption, execution time, and predictive performance while enforcing explicit resource and performance constraints.
\item We integrate the spatial characteristics of the agricultural deployment into the framework to determine the required federation size.
\item We employ constrained Bayesian Optimization to jointly select the DL architecture, aggregation strategy, and number of communication rounds without exhaustively evaluating the configuration space.
\item We evaluate the proposed framework on a plant disease classification task in a resource-constrained IoT setting, demonstrating near-optimal configuration with a limited number of evaluations.
\end{itemize}

The remainder of this paper is organized as follows. Section \ref{sec:Back} provides the background on Federated Learning and the considered deep learning architectures. Section \ref{sec:optimization} formulates the resource-aware HFL configuration problem, while Section \ref{sec:bayesian} presents the proposed constrained Bayesian Optimization approach. Section \ref{sec:evaluation} presents the experimental evaluation. Finally, Section \ref{sec:concl} concludes the paper.

\section{Background}\label{sec:Back}
\subsection{Federated Learning}
In Federated Learning (FL), every participating device uses its own private data to train a local model. A centralized controller collects and combines these individual contributions into a single global model and distributes it back to the devices \cite{mcmahan2017fedavg}. In contrast to conventional single-server FL, HFL introduces intermediate aggregation layers between participating devices and the global server, enabling model aggregation closer to the data sources \cite{9148862}. The aggregation strategy directly affects convergence and model quality. This work considers three strategies: \textbf{(i)} \textit{FedAvg} \cite{mcmahan2017fedavg}, which computes a  weighted average of local parameters; \textbf{(ii)} \textit{FedProx} \cite{li2020fedprox}, which adds a proximal regularization term to limit client drift; and \textbf{(iii)} \textit{FedAvgM} \cite{hsu2019fedavgm}, which introduces server-side momentum to smooth successive global updates.

\subsection{Deep Neural Network Models}
In this work, we consider the following models: \textbf{(i)} \textit{EfficientNet-B0} \cite{tan2019efficientnet}; \textbf{(ii)} \textit{ResNet-50}~\cite{he2016resnet}; and \textbf{(iii)} \textit{MobileNetV3-Large} \cite{howard2019mobilenetv3}. These architectures exhibit different computational characteristics and provide a diverse set of backbone alternatives for evaluating the trade-offs between predictive performance and resource requirements. They therefore define the model-architecture dimension of the optimization search space.

\section{Resource-Aware HFL Configuration Problem}
\label{sec:optimization}

The deployment of HFL systems in resource-constrained IoT environments requires balancing predictive performance against energy consumption and execution time. To address this trade-off, we formulate the HFL configuration as a constrained optimization problem, where deployment requirements define the feasible configuration~space.

The framework incorporates user-defined preferences for energy consumption, execution time, and predictive performance, together with explicit resource and performance constraints. The federation size is determined by the spatial coverage requirements of the agricultural deployment, while the backbone architecture, aggregation strategy, and number of communication rounds constitute the optimization variables. These elements jointly define the resource-aware HFL configuration problem considered in the following sections.

\subsection{Search Space and Deployment Model}

A candidate HFL configuration is represented as

\begin{equation}
\mathbf{x}=(m,a,R),
\qquad
\mathcal{X}
=
\mathcal{M}
\times
\mathcal{A}
\times
\{1,\ldots,R_{\max}\},
\label{eq:searchspace}
\end{equation}

where $m\in\mathcal{M}$ denotes the selected backbone model, $a\in\mathcal{A}$ denotes the aggregation strategy, and $R$ denotes the number of communication rounds. 

The number of participating devices is determined before optimization according to the farm geometry. Let $A_{\mathrm{farm}}$ denote the farm area and $r_d$ the effective sensing or communication radius of one device. Since circular coverage regions cannot cover an arbitrary agricultural area without overlap and boundary losses, a coverage-efficiency coefficient $\rho\in(0,1]$ is introduced. The required federation size is then estimated as

\begin{equation}
N
=
\left\lceil
\frac{A_{\mathrm{farm}}}
{\rho\pi r_d^2}
\right\rceil,
\qquad
\rho=0.8,
\label{eq:numberDevices}
\end{equation}

where the adopted value assumes that $80\%$ of the ideal circular region contributes to effective farm coverage. The value of $\rho$ can be adjusted according to the geometry and coverage characteristics of a specific deployment \cite{ammari}. Consequently, $N$ is treated as a deployment parameter rather than a variable selected by the Bayesian optimizer.

Let $N_0$ and $R_0$ denote the federation size and number of communication rounds used in the reference experiments. For each model--aggregator pair $(m,a)$, the measured energy consumption $E_0(m,a)$ and execution time $T_0(m,a)$ are extended to different deployment conditions using first-order scaling approximations. Motivated by resource-aware FL models, which characterize the overall computation and communication cost as accumulating across participating devices and communication rounds \cite{wang2019adaptive, yang2021energy}, the energy consumption is approximated as

\begin{equation}
E(\mathbf{x},N)
=
E_0(m,a)
\left(\frac{N}{N_0}\right)
\left(\frac{R}{R_0}\right),
\label{eq:energyScaling}
\end{equation}

and

\begin{equation}
T(\mathbf{x})
=
T_0(m,a)
\left(\frac{R}{R_0}\right).
\label{eq:timeScaling}
\end{equation}

The energy model accounts for changes in both federation size and number of communication rounds. The execution-time model depends only on $R$, under the assumption that local client operations are executed predominantly in parallel and that additional coordination overhead remains limited.

To characterize the dependence of predictive performance on the number of communication rounds, a saturation function is fitted separately for each model--aggregator pair:

\begin{equation}
F_1(\mathbf{x})
=
F_{1,0}(m,a)
+
\left[
F_{1,\infty}(m,a)-F_{1,0}(m,a)
\right]
\left(1-e^{-k_{m,a}R}\right),
\label{eq:f1Rounds}
\end{equation}

where $F_{1,0}(m,a)$ denotes the initial performance, $F_{1,\infty}(m,a)$ the expected plateau, and $k_{m,a}$ the corresponding convergence rate. These parameters are estimated from the round-level validation F1-score measurements recorded for each configuration. In the present experimental evaluation, however, direct measured F1 values are used whenever they are available. The fitted function is retained as a general round-dependent approximation for configurations or communication rounds for which direct measurements are not available.

The upper search limit $R_{\max}$ is determined from the experimentally available communication-round range. Although the fitted convergence curves can also be used to inspect the expected plateau behavior beyond the observed interval, round-level measurements in the present study are available for $R=1,\ldots,30$. The experimental search space is therefore restricted to $R_{\max}=30$, avoiding reliance on extrapolated performance values during the evaluation of the optimization method.

\subsection{Weighted Objective and Constraints}

Energy consumption and execution time are expressed in different physical units and may vary substantially across deployment conditions. They are therefore normalized directly by the budgets specified by the user, while the F1-score requires no additional scaling because it already lies within $[0,1]$:

\begin{equation}
\hat{E}(\mathbf{x},N)
=
\frac{E(\mathbf{x},N)}{E_{\mathrm{budget}}},
\quad
\hat{T}(\mathbf{x})
=
\frac{T(\mathbf{x})}{T_{\mathrm{budget}}},
\quad
\hat{F}_1(\mathbf{x})
=
F_1(\mathbf{x}).
\label{eq:normalization}
\end{equation}

This formulation provides a direct interpretation of resource usage, since values of $\hat{E}$ or $\hat{T}$ greater than one indicate that the corresponding deployment budget has been exceeded.

The normalized quantities are combined into the scalar objective

\begin{equation}
\mathcal{L}(\mathbf{x})
=
\lambda_1\hat{E}(\mathbf{x},N)
+
\lambda_2\hat{T}(\mathbf{x})
+
\lambda_3
\left[
1-\hat{F}_1(\mathbf{x})
\right],
\label{eq:loss}
\end{equation}

where the user-defined coefficients $\lambda_1$, $\lambda_2$, and $\lambda_3$ express the relative importance assigned to energy consumption, execution time, and predictive performance, respectively. To form a valid convex combination, they satisfy

\begin{equation}
\lambda_i\geq0,
\qquad
\sum_{i=1}^{3}\lambda_i=1.
\label{eq:lambdaConditions}
\end{equation}

The optimization is additionally restricted by the maximum allowable energy consumption and execution time, the minimum required predictive performance, and the admissible number of communication rounds:

\begin{equation}
\begin{aligned}
E(\mathbf{x},N) &\leq E_{\mathrm{budget}}, \\
T(\mathbf{x}) &\leq T_{\mathrm{budget}}, \\
F_1(\mathbf{x}) &\geq F_{1,\mathrm{required}}, \\
1 &\leq R\leq R_{\max}.
\end{aligned}
\label{eq:constraints}
\end{equation}

The budget-normalized terms determine the relative cost of feasible candidates, while the explicit constraints exclude configurations that violate the deployment requirements.

\subsection{Optimization Problem}

Let $\Omega\subseteq\mathcal{X}$ denote the subset of configurations satisfying the constraints in (\ref{eq:constraints}). The configuration-selection problem is then expressed as

\begin{equation}
\mathbf{x}^{*}
=
\arg\min_{\mathbf{x}\in\Omega}
\mathcal{L}(\mathbf{x}).
\label{eq:finalProblem}
\end{equation}

For a previously untested configuration, evaluating $\mathcal{L}(\mathbf{x})$ requires executing the corresponding HFL training process and obtaining its predictive and resource-related quantities. The configuration-selection problem can therefore be treated as an expensive black-box optimization task. In the present study, previously collected round-level measurements are used to emulate these expensive evaluations, allowing the proposed optimizer to be assessed under a controlled setting and directly compared with the optimum obtained from the complete discrete search space. The constrained Bayesian Optimization procedure used to perform this search is presented in the following section.

\section{Constrained Bayesian Optimization}
\label{sec:bayesian}

Although the optimization problem in (\ref{eq:finalProblem}) is defined over a finite search space, exhaustively evaluating every candidate becomes increasingly expensive as additional architectures, aggregation strategies, communication-round settings, or deployment conditions are introduced. Each previously untested configuration may require the execution of the corresponding HFL training process together with the collection of predictive and resource-related measurements. Bayesian Optimization (BO) is therefore employed to guide the search toward promising feasible configurations while limiting the number of expensive evaluations \cite{snoek2012practical}.

The proposed procedure treats objective quality and constraint satisfaction separately. A Gaussian Process regression model approximates the scalar objective $\mathcal{L}(\mathbf{x})$, while a second probabilistic model estimates the likelihood that a candidate satisfies the deployment constraints. The two models are combined through a constrained acquisition function that favors configurations expected to improve the current best feasible solution while maintaining a high probability of feasibility.

\subsection{Surrogate Models and Candidate Encoding}

The search variables contain both categorical and numerical components. For a candidate
$\mathbf{x}=(m,a,R)$, the backbone model $m$ and aggregation strategy $a$ are represented through one-hot encoding, while the communication round is normalized to the interval $[0,1]$ as

\begin{equation}
\tilde{R}
=
\frac{R-1}{R_{\max}-1}.
\label{eq:roundNormalization}
\end{equation}

The resulting numerical representation is

\begin{equation}
\mathbf{z}(\mathbf{x})
=
[
\mathbf{e}_{m},
\mathbf{e}_{a},
\tilde{R}
],
\label{eq:boEncoding}
\end{equation}

where $\mathbf{e}_{m}$ and $\mathbf{e}_{a}$ denote the one-hot vectors associated with the selected model and aggregation strategy, respectively.

Given the set of already evaluated configurations
$\mathcal{D}_t=\{(\mathbf{z}_i,\mathcal{L}_i)\}_{i=1}^{t}$,
a Gaussian Process regression model is fitted to the observed objective values:

\begin{equation}
\mathcal{L}(\mathbf{z})
\sim
\mathcal{GP}
\left(
\mu(\mathbf{z}),
k(\mathbf{z},\mathbf{z}')
\right),
\label{eq:gpObjective}
\end{equation}

providing, for each unevaluated candidate, a predictive mean
$\mu_t(\mathbf{z})$ and standard deviation
$\sigma_t(\mathbf{z})$. In the implementation, a Mat\'ern kernel is used for the objective surrogate.

Constraint satisfaction is modeled independently. Each evaluated candidate is assigned the binary label

\begin{equation}
y_i
=
\begin{cases}
1, & \mathbf{x}_i\in\Omega,\\
0, & \mathbf{x}_i\notin\Omega,
\end{cases}
\label{eq:feasibilityLabel}
\end{equation}

where $\Omega$ is the feasible set defined in (\ref{eq:constraints}). A Gaussian Process classifier is then trained on these labels to estimate

\begin{equation}
p_t(\mathbf{x})
=
P(\mathbf{x}\in\Omega\mid\mathcal{D}_t),
\label{eq:feasibilityProbability}
\end{equation}

which represents the probability that a candidate satisfies the energy, execution-time, and predictive-performance requirements.

\subsection{Acquisition and Search Procedure}

The optimization begins with a small set of unique randomly selected configurations. After these initial evaluations, the objective surrogate and feasibility model are updated using all observations collected so far. If no feasible configuration has yet been observed, candidate selection is driven by the estimated probability of feasibility until the first feasible solution is identified. Candidate selection is based on Expected Improvement (EI), which quantifies the expected reduction relative to the best feasible objective value observed at iteration~$t$.

For minimization, EI is computed as

\begin{equation}
\mathrm{EI}_t(\mathbf{x})
=
\Delta_t(\mathbf{x})
\Phi\!\left(
\frac{\Delta_t(\mathbf{x})}
{\sigma_t(\mathbf{x})}
\right)
+
\sigma_t(\mathbf{x})
\phi\!\left(
\frac{\Delta_t(\mathbf{x})}
{\sigma_t(\mathbf{x})}
\right),
\label{eq:expectedImprovement}
\end{equation}

where

\begin{equation}
\Delta_t(\mathbf{x})
=
\mathcal{L}_{\mathrm{best}}
-
\mu_t(\mathbf{x})
-
\xi,
\label{eq:improvement}
\end{equation}

$\Phi(\cdot)$ and $\phi(\cdot)$ denote the standard normal cumulative and probability density functions, respectively, and $\xi$ controls the exploration--exploitation trade-off.

To account explicitly for the deployment constraints, EI is weighted by the estimated probability of feasibility, following the constrained BO formulation in \cite{gardner14}:

\begin{equation}
\alpha_t(\mathbf{x})
=
\mathrm{EI}_t(\mathbf{x})
\,
p_t(\mathbf{x}).
\label{eq:constrainedEI}
\end{equation}

At each iteration, the acquisition function is evaluated over the set of configurations that have not yet been tested, and the next candidate is selected according to

\begin{equation}
\mathbf{x}_{t+1}
=
\arg\max_{\mathbf{x}\in
\mathcal{X}\setminus\mathcal{D}_t}
\alpha_t(\mathbf{x}).
\label{eq:nextCandidate}
\end{equation}

Only the selected candidate is then evaluated using the actual HFL objective and constraints, after which both probabilistic models are updated and the process is repeated until the predefined evaluation budget is reached. Restricting acquisition evaluation to previously unseen candidates also guarantees that no configuration is evaluated more than once. In the present study, six unique configurations are used for initialization and the overall optimization budget is limited to 30 evaluations.

\section{Experimental Evaluation}
\label{sec:evaluation}

\subsection{Optimization Setup and Ground-Truth Benchmark}

The proposed optimization framework was evaluated using the nine HFL model--aggregator combinations considered in the previous experiments, with the number of communication rounds restricted to $R\in\{1,\ldots,30\}$. The resulting discrete search space contains

\begin{equation}
|\mathcal{X}| = 3 \times 3 \times 30 = 270
\end{equation}

candidate configurations. The reference measurements correspond to $N_0=10$ participating clients and $R_0=30$ communication rounds. For the deployment scenario considered in the optimization experiments, we assume a farm area of $A_{\mathrm{farm}}=10\,000~\mathrm{m}^2$ and an effective device coverage radius of $r_d=20~\mathrm{m}$. With $\rho=0.8$, the spatial coverage model in (\ref{eq:numberDevices}) results in $N=10$, matching the federation size used in the reference experiments. Consequently, no additional scaling with respect to the number of participating clients is introduced in the present evaluation, while the proposed formulation remains applicable to deployments with different federation sizes.

The optimization parameters were set to

\begin{equation}
E_{\mathrm{budget}}=20~\mathrm{Wh},
\quad
T_{\mathrm{budget}}=400~\mathrm{s},
\quad
F_{1,\mathrm{required}}=0.80,
\end{equation}

with objective weights

\begin{equation}
(\lambda_1,\lambda_2,\lambda_3)
=
(0.4,0.2,0.4).
\end{equation}

\begin{figure*}[!t]
    \centering
    \includegraphics[width=0.75\textwidth]{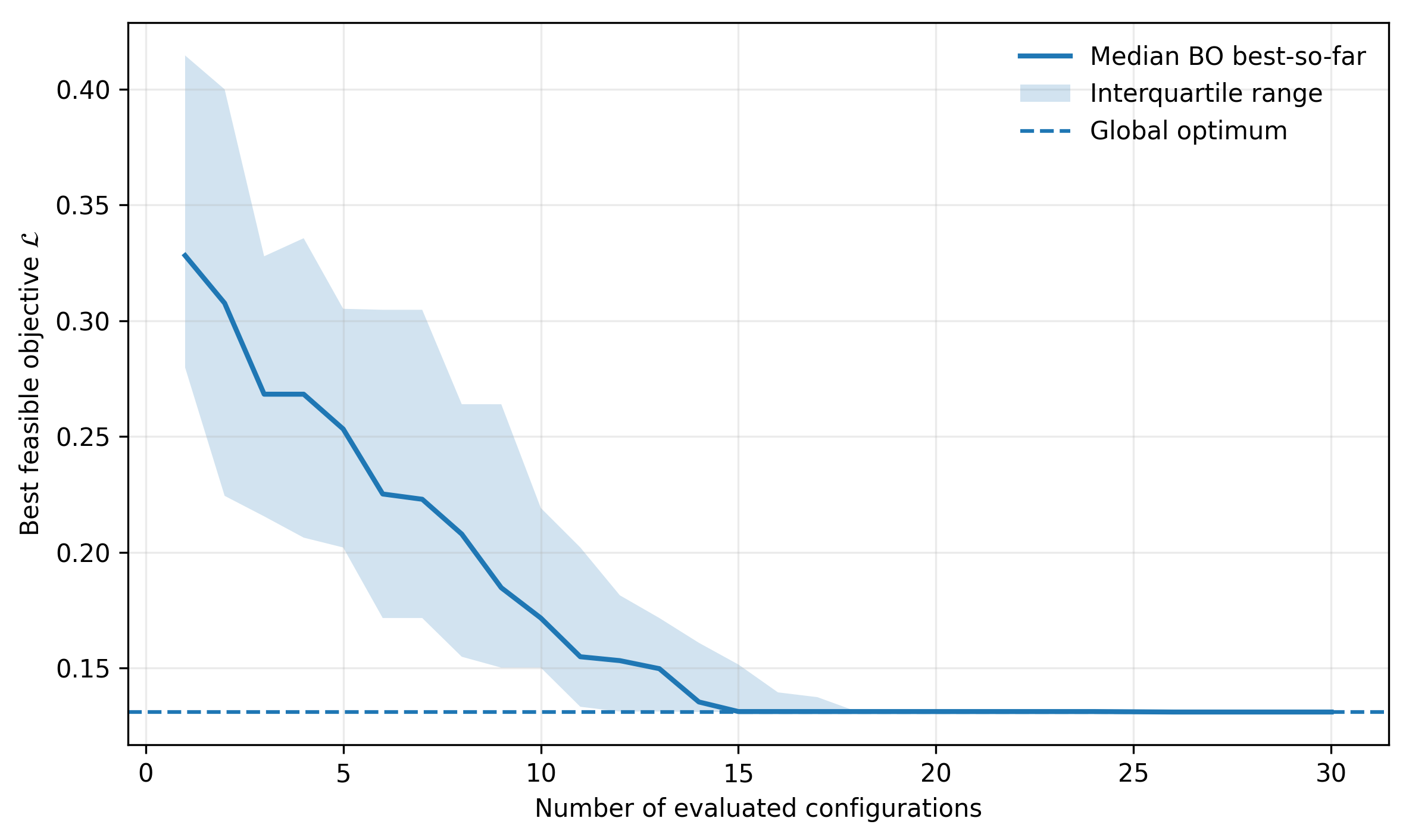}
    \caption{Convergence of constrained Bayesian Optimization over 30 independent runs. The solid curve represents the median best feasible objective observed up to each evaluation, the shaded region denotes the interquartile range, and the dashed line indicates the global optimum obtained from exhaustive evaluation of the complete search space.}
    \label{fig:boConvergence}
\end{figure*}

Energy consumption and execution time at intermediate communication rounds were obtained through (\ref{eq:energyScaling}) and (\ref{eq:timeScaling}), while the directly measured validation F1-score was used for each available round. Since measurements were available for the complete search space, an exhaustive evaluation of all 270 candidates was performed once as an offline reference. This exhaustive search is not part of the proposed optimization procedure, but is used exclusively to establish the true optimum and quantify the performance of Bayesian Optimization.

Of the 270 candidate configurations, 141 satisfied all deployment constraints. The globally optimal feasible solution was EfficientNet-B0 with FedAvg at $R=4$, yielding

\begin{equation}
E=1.556~\mathrm{Wh},
\qquad
T=40.156~\mathrm{s},
\qquad
F_1=0.800326,
\end{equation}

with an objective value of

\begin{equation}
\mathcal{L}^{*}=0.131068.
\end{equation}

The second-best configuration was ResNet-50 with FedProx at $R=2$, with $\mathcal{L}=0.131266$, corresponding to a relative difference of only $0.152\%$ from the global optimum.

\subsection{Bayesian Optimization Results}

To evaluate robustness with respect to initialization, the constrained Bayesian Optimization procedure was repeated for 30 independent random seeds. Each run was initialized with six unique randomly selected configurations and was limited to a total budget of 30 evaluations. Therefore, each optimization run evaluated only

\begin{equation}
\frac{30}{270}\times100 = 11.11\%
\end{equation}

of the complete configuration space.

\begin{table}[h]
\centering
\caption{Performance of constrained Bayesian Optimization over 30 independent runs.}
\label{tab:boResults}
\begin{tabular}{lc}
\hline
Metric & Result \\
\hline
Search-space size & 270 \\
Evaluations per run & 30 (11.11\%) \\
Feasible-run rate & 100\% \\
Exact global optimum & 63.33\% \\
Within 1\% of optimum & 100\% \\
Mean optimality gap & 0.056\% \\
Median optimality gap & 0.000\% \\
Maximum optimality gap & 0.152\% \\
Median best iteration & 15.5 \\
\hline
\end{tabular}
\end{table}

Table~\ref{tab:boResults} summarizes the final optimization performance over all 30 runs. A feasible solution was identified in every run, while the exact global optimum was recovered in $63.33\%$ of the cases. More importantly, every run terminated with a solution within $1\%$ of the exhaustive-search optimum. The mean relative optimality gap was only $0.056\%$, the median gap was $0\%$, and the maximum observed gap was $0.152\%$. In all runs in which the exact optimum was not selected, the final solution corresponded to the second-best configuration identified by exhaustive evaluation.

The convergence behavior is shown in Fig.~\ref{fig:boConvergence}. The median best-so-far objective decreases rapidly as additional configurations are evaluated and approaches the exhaustive-search optimum after approximately 15--20 evaluations. After only 15 evaluations, corresponding to $5.56\%$ of the complete search space, the median optimality gap was already $0.152\%$, while $56.67\%$ of the runs had reached a solution within $1\%$ of the optimum. This proportion increased to $80.00\%$ after 20 evaluations, $86.67\%$ after 25 evaluations, and $100\%$ at the final budget of 30 evaluations.

The final best solution of each run was first identified at a median iteration of 15.5, corresponding to approximately $5.74\%$ of the complete search space. These results indicate that the proposed constrained Bayesian Optimization procedure can reliably concentrate the search around the globally optimal region while requiring only a small fraction of the evaluations needed by exhaustive search.

\section{Conclusions}\label{sec:concl}
This paper presented a constrained Bayesian Optimization framework for efficient HFL configuration in resource-constrained IoT environments for plant disease classification. The framework jointly optimizes the learning architecture, aggregation strategy, and communication rounds under energy, execution-time, and predictive-performance constraints. Experimental results show that only 11.11\% of the search space is explored, while all runs identified solutions within 1\% of the exhaustive-search optimum. These results demonstrate the potential of constrained Bayesian Optimization for efficient and resource-aware HFL deployment in smart agricultural IoT environments.

\section*{Acknowledgment}
This work was funded under the COIN-3D project, which has received funding from the European Union’s Horizon Europe research and innovation programme under grant agreement No. 101159667.

\balance
\bibliographystyle{IEEEtran}
\bibliography{references.bib}
\end{document}